\documentclass[10pt,letterpaper]{article}

\usepackage{cogsci}
\cogscifinalcopy 

\usepackage{pslatex}
\usepackage{apacite} 
\usepackage[round]{natbib} 
\usepackage{graphicx} 
\usepackage{subcaption} 
\usepackage[font=small,labelfont=bf]{caption} 
\usepackage{amsmath} 
\usepackage{amssymb} 
\renewcommand\bibliographytypesize{\scriptsize} 

\usepackage{soul} 
\usepackage{color}
\usepackage{xcolor,colortbl}

\newcommand{\bcaption}[2]{\caption{\textbf{#1} #2}}

\title{Generating new concepts with hybrid neuro-symbolic models}
 
\author{{\large \bf Reuben Feinman (reuben.feinman@nyu.edu)} \\
  Center for Neural Science \\
  New York University
  \And {\large \bf Brenden M. Lake (brenden@nyu.edu)} \\
  Department of Psychology and Center for Data Science \\
  New York University}

\begin{document}

\maketitle

\begin{abstract}

Human conceptual knowledge supports the ability to generate novel yet highly structured concepts, and the form of this conceptual knowledge is of great interest to cognitive scientists.
One tradition has emphasized structured knowledge, viewing concepts as embedded in intuitive theories or organized in complex symbolic knowledge structures.
A second tradition has emphasized statistical knowledge, viewing conceptual knowledge as an emerging from the rich correlational structure captured by training neural networks and other statistical models.
In this paper, we explore a synthesis of these two traditions through a novel neuro-symbolic model for generating new concepts.
Using simple visual concepts as a testbed, we bring together neural networks and symbolic probabilistic programs to learn a generative model of novel handwritten characters.
Two alternative models are explored with more generic neural network architectures.
We compare each of these three models for their likelihoods on held-out character classes and for the quality of their productions, finding that our hybrid model learns the most convincing representation and generalizes further from the training observations.

\end{abstract}

\section{Introduction}

People can synthesize new concepts in imaginative ways; architects design new houses, chefs invent new recipes, and entrepreneurs create new business models.
The resulting productions exhibit novel variations but maintain important structural consistencies with known entities \citep{Ward1994}.
In contrast, state-of-the-art generative models from machine learning struggle with creative imagination, producing samples that either closely mimic the training data or that exhibit anomalous characteristics \citep{Lake2019}.
How do people create novel yet coherent new concepts?
How can we understand these abilities in computational terms?

Human conceptual knowledge plays a central role in creative generalization. 
A chef knows not only a repertoire of recipes, but also understands that recipes are built from \textit{reusable} ingredients (e.g. carrots, flour, butter), and that these ingredients satisfy specific roles (thickening, seasoning, greasing). 
Furthermore, a chef understands which ingredients can substitute for others (e.g. butter for oil when greasing) and which should never be combined (e.g. ketchup and milk). 
In addition, they understand that recipes are composed of reusable causal procedures (cutting, whisking, browning), and they know how to compose these procedures in terms of order and substitutability. 
This \textit{causal} and \textit{compositional} knowledge is essential to understanding a culinary concept, as opposed to merely executing it, and is essential to a chef's ability to create new culinary concepts such as ``carrots tartar" or ``pea guacamole." 

There have been two traditions of work on computational models of conceptual knowledge. 
The first tradition emphasizes ``structured knowledge" for capturing relations between concepts and correlations between conceptual features, viewing concepts as embedded in intuitive theories \citep{Murphy1985} or capturing structured knowledge through symbolic representations such as hierarchies, trees, grammars and programs \citep{Kemp2008,Kemp2009,Tenenbaum2011}.
This tradition has prioritized the compositional and causal nature of conceptual knowledge, as emphasized through accounts of concept learning as program induction \citep{Goodman2008,Stuhlmuller2010,Lake2015,Goodman2015,Ellis2018,LakeFractals2019}.
The Bayesian Program Learning (BPL) framework \citep{Lake2015}, for example, demonstrates how to learn programs from images to express the causal and compositional nature of concepts and background knowledge.
Although these models offer a convincing account for how strong inductive biases support flexible generalization, they often make simplifying and rigid parametric assumptions about the distributions of concepts in pursuit of a structured representation. 
As a result, they so far have been unsuccessful in characterizing the most complex correlations and invariances associated with human concepts in raw, high-dimensional stimulus spaces.

The second tradition in models of conceptual knowledge emphasizes ``statistical knowledge," a more amorphous form of background knowledge that is often not amenable to symbolic description.
In the statistics view, conceptual knowledge manifests as complex systems of patterns and correlations recorded from observations.
The meaning of a word, for example, can be derived from its patterns of co-occurrance with other words \citep{Deerwester1990}.
Similarly, latent representations of objects and other sensory stimuli can be derived from ``suspicious coincidences" noted in the data \citep{Barlow1989}.
The statistics view emphasizes emergence, where conceptual knowledge emerges from the interaction of simpler processes, as operationalized through training neural network architectures \citep{McClelland2010a}. 
Although a powerful modeling tool, standard neural networks do not explicitly model the compositional and causal structure of concepts.
As result, they have difficulty generalizing to examples that vary systematically from training \citep{Marcus2003,LakeBaroni2018}, and to novel tasks, especially those that demand more generative and creative abilities \citep{Lake2016,Lake2019}.

Our goal in this paper is to explore generative models of concepts at the interface of these structured and statistical traditions, with the aim of combining strengths from both approaches. Previous efforts to integrate these traditions have demonstrated ways of performing statistical inference over structured representations \citep{Tenenbaum2011}. This includes models of concept learning as Bayesian inference over fully-symbolic expressions in formal logical \citep{Goodman2008,Piantadosi2016c}, or models of inductive reasoning supported by structured intuitive theories \citep{Kemp2009}. In accounts of this nature, statistics is primary in selecting between structured symbolic hypotheses \citep{Kemp2008,Perfors2011,Lake2015,LakeFractals2019}, but plays little role in forming the individual hypotheses themselves. Specifically, each hypothesis may only have a few parametric distributions that need to be inferred (Gaussians, multinomials, etc.), if any.

Here we aim to more thoroughly integrate the structured and statistical traditions through hybrid neuro-symbolic generative models.
Our goal is to devise a causal generative model with explicit compositional structure, and with complex correlations represented implicitly through neural networks rather than simple parametric distributions. 
We use simple visual concepts -- handwritten characters from the world's languages -- as a case study for exploring neuro-symbolic models of concept generation. The Omniglot dataset \citep{Lake2015} of handwritten characters provides an excellent preliminary modeling environment: it contains a large number of natural, simple concepts that people learn and use, and it has been explored extensively in prior work from both cognitive science and AI. 
Following the mixture density network framework for handwriting generation \citep{Graves2013}, we explore three distinct generative neural architectures, varying the strength and form of inductive bias imposed on the model, including their position on the neuro-symbolic spectrum and the fidelity in which compositionality and causality are presented.
We evaluate the generalization capacity of these models by comparing their log-likelihoods on a holdout set of characters. 
Furthermore, we analyze the samples produced by each model, looking for characters that are qualitatively consistent but sufficiently dissimilar from the training set. 
We find that a hybrid neuro-symbolic architecture with the strongest form of compositional structure exhibits the best generalization performance, and that it generates characters that are highly consistent with human drawings. 
In contrast, the generic neural models exhibit weaker performance on the holdout set, and they produce characters that more closely mimic the training examples.

\section{Related Work}
In the machine learning community, there have been a number of works studying generative neural network models for handwritten characters, including DRAW \citep{Gregor2015}, AIR \citep{Eslami2016} and SPIRAL \citep{Ganin2018}. 
Although these models learn a procedure to generate new characters, they do not use the human drawing data from Omniglot, and therefore the generative process may not reflect the true causal processes of human character production. 
Our goal is different in that we aim to model the causal process of human handwriting directly from drawing data.

\begin{figure*}[t]
    \centering
    \includegraphics[width=0.9\hsize]{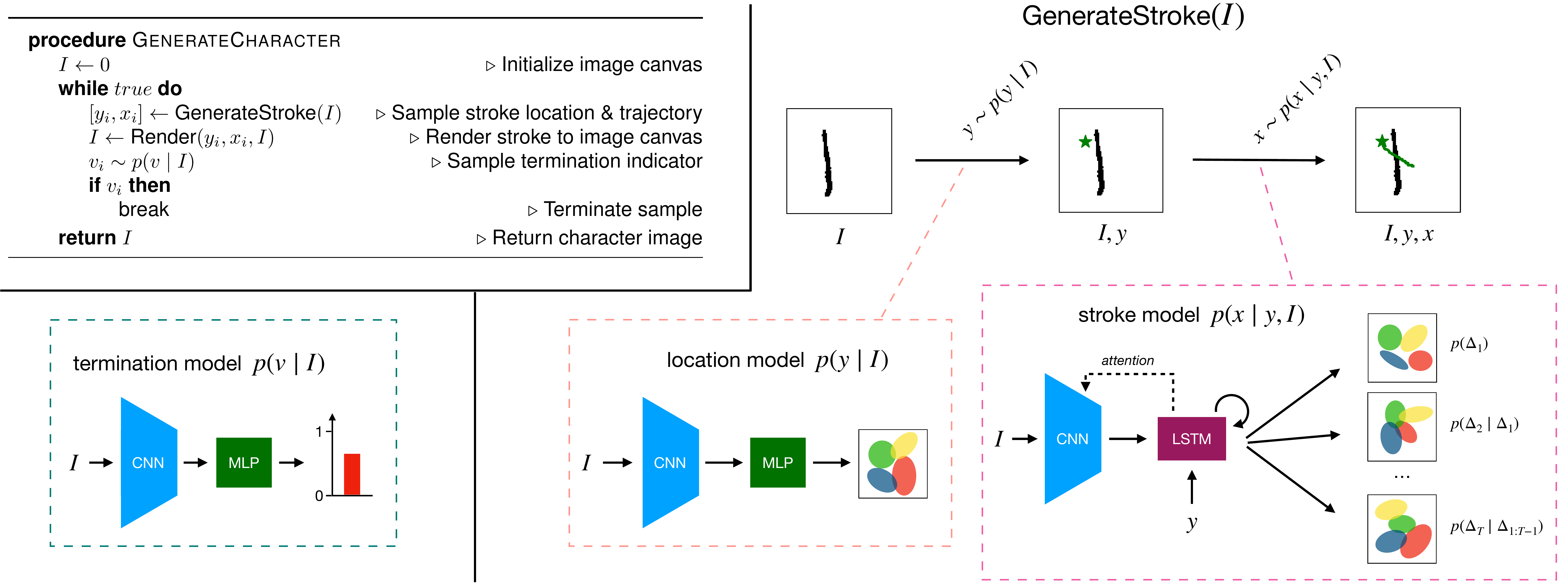}
	\bcaption{Full neuro-symbolic (Full NS) model.}{
	Our Full NS model produces character samples one stroke at a time.
	The procedure \texttt{GenerateCharacter} consists of sequentially reading from and rendering to an \textit{image canvas}, which is initialized to zero. 
	At each time step, the current canvas $I$ is fed to procedure \texttt{GenerateStroke}, which produces a stroke sample.
	The canvas is first processed by the \textit{location model}, a CNN-MLP architecture that processes the image and returns a Gaussian mixture model (GMM) distribution for the starting location of the next stroke $y$. 
	The location $y$ is then sampled and passed along with $I$ to the \textit{stroke model}. 
	The stroke model processes $I$ with a CNN and feeds the embedding to an LSTM with attention. 
	The LSTM samples a stroke trajectory $x$ sequentially one offset at a time using GMM outputs.
	The sampled stroke is passed to a symbolic renderer, and the updated image canvas is then processed by a \textit{termination model} that decides whether to continue the character sample.
	}
    \label{fig:arch_canvas}
\end{figure*}

\cite{Ha2018} introduced a neural network architecture called Sketch-RNN to model human drawing data for simple objects like cats, firetrucks, and windmills. 
Although their goal loosely resembles our own, the Sketch-RNN model is trained on just a single class of objects at one time (e.g. ``cat"), and it receives 70,000 examples from the class.
In contrast, our motivation is to model human conceptual knowledge of handwriting concepts in general. 
This background knowledge plays a central role in creative generalization, enabling people to synthesize new concepts that deviate from the observed entities.
We train our models on many character classes at once, providing only 20 training examples of each class and asking them to generate new character concepts.
The Sketch-RNN model has not been applied in this way.

Most related to our work is the Bayesian Program Learning (BPL) approach of  \cite{Lake2015} that was also applied to the simple visual concepts in Omniglot.
BPL is a parametric Bayesian model that captures causal, compositional structure in human background knowledge of handwriting, and shows that these ingredients are important for few-shot learning of new character concepts. 
Beyond supporting few-shot learning, the BPL character prior can also generate new character concepts by unconditional sampling. 
Although a powerful demonstration of compositional representation, the BPL parametric model makes many simplifying assumptions about characters.
For example, it assumes that strokes in a character are generated largely independently from each other in the prior (although they are strongly correlated in the posterior).
As result, new characters generated by the model often lack the rich correlation structure of human drawings.
We build on this work and develop a new neuro-symbolic model that represents the compositional structure of characters while using neural networks to capture richer correlations.


\section{Omniglot Case Study}
We use simple visual concepts as a case study for modeling conceptual structure. The Omniglot dataset contains human drawings of characters from 50 unique alphabets, providing a large set of cognitively natural concepts that are simple enough for evaluating models \citep{Lake2015,Lake2019}.
In our experiments, we use drawings from the Omniglot background set to train our models, which contains 30 alphabets and a total of 19,280 unique drawings. 
We also use 10 alphabets from the Omniglot evaluation set as a holdout set for quantitative evaluations, reserving the remaining 10 alphabets for future work on few-shot classification.

\begin{figure}[h]
    \centering
    \includegraphics[width=0.5\hsize]{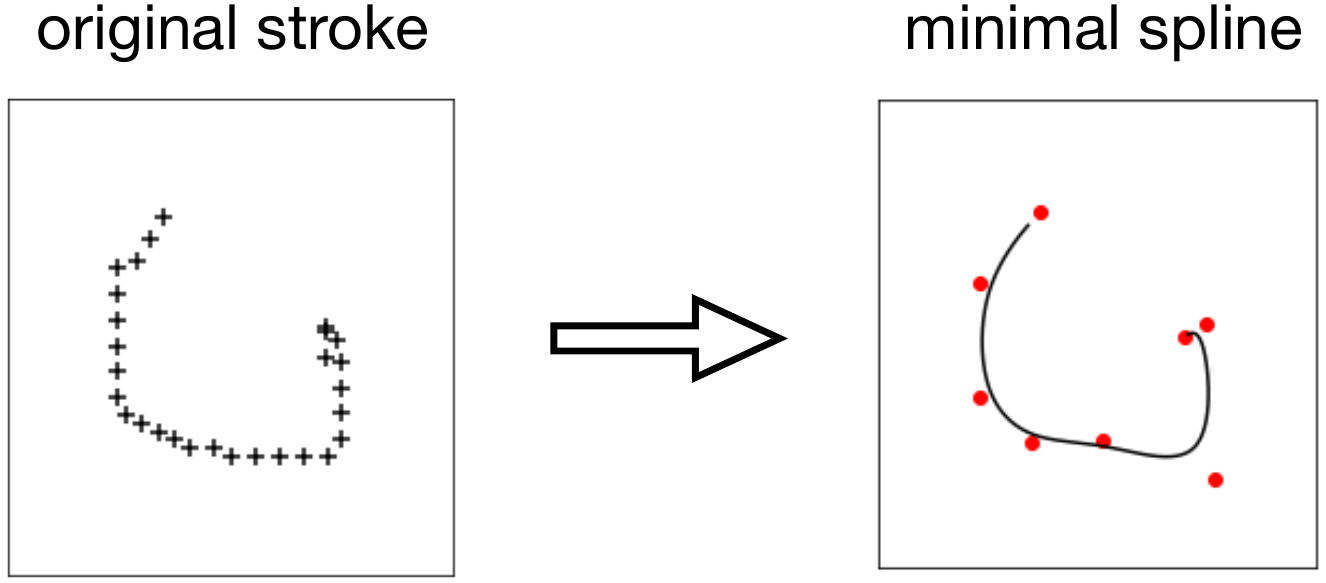}
	\bcaption{Spline representation.}{Raw strokes (left) are converted into minimal splines (right) using least-squares optimization. Crosses (left) indicate pen locations and red dots (right) indicate spline control points.}
    \label{fig:spline_format}
\end{figure}

In the drawing data, a stroke is represented as a variable-length sequence of pen locations $\{z_1,...,z_T\}$, with $z_i \in \mathbb{R}^2$ (Fig. \ref{fig:spline_format}, left).
During pre-processing, we convert each stroke into a minimal spline representation using least-squares optimization (Fig. \ref{fig:spline_format}, right), borrowing the B-spline tools from \cite{Lake2015}. 
The number of spline control points depends on the stroke complexity and is determined by a residual threshold.
Furthermore, we removed small strokes using a threshold on the trajectory length. 
These processing steps help suppress noise and emphasize signal in the drawings. 
Our generative models are trained to produce character drawings, where each drawing is represented as an ordered set of splines (strokes).
The number of strokes, and the number of spline coordinates per stroke, are allowed to vary.

\section{Neuro-Symbolic Model}
Our primary interest is to test whether a hybrid neuro-symbolic model can capture the compositional, causal structure in a large corpus of simple natural concepts. 
The architecture and sampling procedure of our hybrid model, which we call the ``Full Neuro-Symbolic" (Full NS) model, is given in Fig. \ref{fig:arch_canvas}.
Compared to generic neural networks, the Full NS model lies closer to \textit{structure} on the structure-statistics spectrum, possessing a much stronger inductive bias.
As in BPL \citep{Lake2015}, the generative model is a probabilistic program that captures real compositional and causal structure by sampling characters as a sequence of parts and locations/relations. 
Unlike BPL, the model has a symbolic engine that renders each part to an \textit{image canvas} before producing the next one, and parts are generated using a powerful recurrent neural network that encodes and attends to the current canvas.
Although correlations between parts can be captured through a process of rendering and then encoding, the model does not allow arbitrary information to flow between parts and variables as in monolithic neural networks.

\begin{figure*}[t]
    \centering
    \includegraphics[width=0.95\hsize]{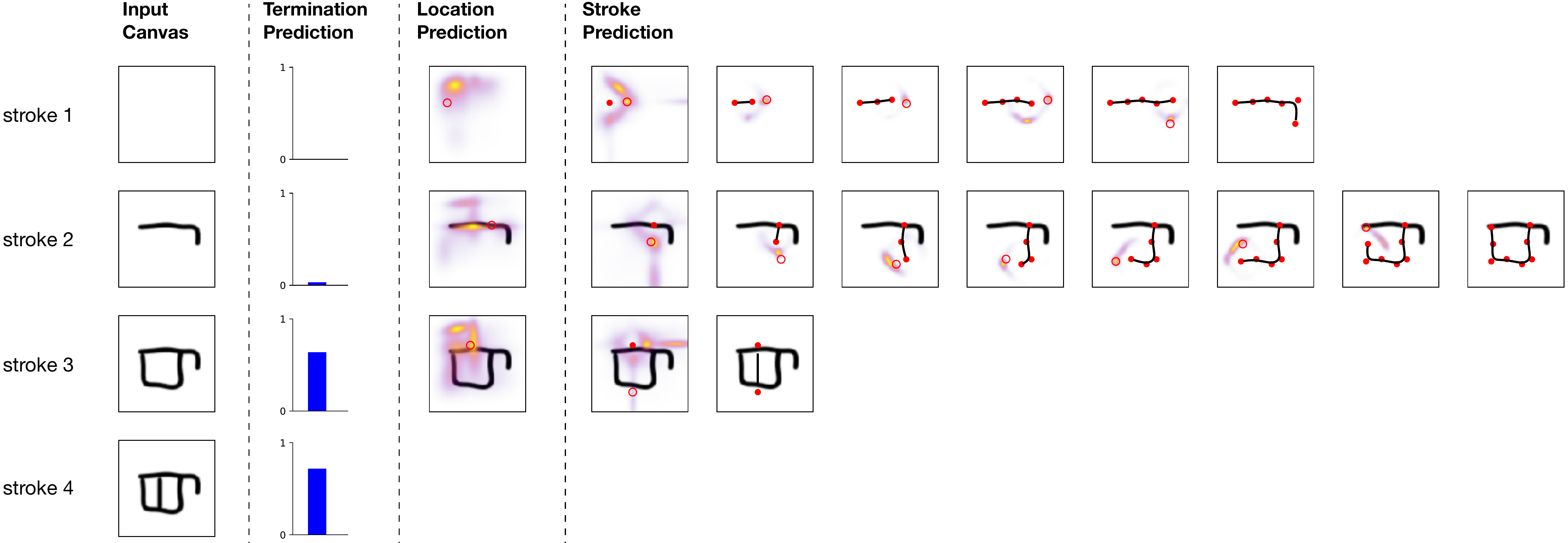}
	\bcaption{Predictions of the Full NS model for a test character.}{
	After each stroke, the model receives the current image canvas (``Input Canvas") and makes a series of predictions. \textbf{Termination Prediction.} First, the model predicts a termination probability $p$ (blue bar), i.e. a probability of terminating the drawing. \textbf{Location Prediction.} Next, the model predicts a probability density for the next stroke's starting location. The heatmap indicates the predicted density, and the hollow red dot indicates the ground-truth location. \textbf{Stroke Prediction.} Finally, the model predicts an auto-regressive probability density for the next stroke's trajectory (the ``stroke"). Red dots indicate the previous control points, heatmaps indicate the predicted density for the next control point, and hollow red dot indicates the ground-truth next control point.
	}
    \label{fig:model_predictions}
\end{figure*}

The Full NS model represents a character as a sequence of strokes, with each stroke decomposed into a starting location $y_t \in \mathbb{R}^2$, conveying the first spline control point, and a stroke trajectory $x_t = \{\Delta_1, ..., \Delta_N \}$, conveying deltas between spline control points.
It generates characters one stroke at a time, using a symbolic rendering procedure called \texttt{Render}, as an intermediate processing step after forming each stroke.
An image canvas $I$ is used as a memory state to convey information about previous strokes. 
At each time step $t$, the next stroke's starting location and trajectory are sampled with procedure \texttt{GenerateStroke}.
In this procedure, the current image canvas $I$ is first read by the \textit{location model} (Fig. \ref{fig:arch_canvas}; bottom middle), a convolutional neural network (CNN) that processes the image and returns a probability distribution for starting location $y_t$:
\begin{align*}
    \quad y_t \sim p(y_t \mid I).
\end{align*}
A visualization of the density $p(y_t \mid I)$ is given in Fig. \ref{fig:model_predictions}, ``Location Prediction."
The starting location $y_t$ is then passed along with the image canvas $I$ to the \textit{stroke model} (Fig. \ref{fig:arch_canvas}; bottom right), a Long Short-Term Memory (LSTM) architecture with a CNN-based image attention mechanism inspired by \cite{Xu2016}. 
The stroke model samples the next stroke trajectory $x_t$ sequentially one offset at a time, selectively attending to different parts of the image canvas at each sample step and combining this information with the context of $y_t$:
\begin{align*}
    \quad x_t \sim p(x_t \mid y_t, I).
\end{align*}
A visualization of the auto-regressive density $p(x_t \mid y_t, I)$ is given in Fig. \ref{fig:model_predictions}, ``Stroke Prediction." Finally, a similar network decides when to terminate the character, $p(v_t \mid I)$.

\subsubsection{Mixture Outputs.}
Both our location model and stroke model follow a technique from \cite{Graves2013}, who proposed to use neural networks with mixture outputs to model handwriting data.
The parameters $\theta = \{\pi^{1:K}, \mu^{1:K}, \sigma^{1:K}, \rho^{1:K}\}$ output by our network specify a Gaussian mixture model (GMM) with K components (Fig. \ref{fig:arch_canvas}; colored ellipsoids), where $\pi^k \in (0,1)$ is the mixture weight of the $k^{\text{th}}$ component, $\mu^k \in \mathbb{R}^2$ its means, $\sigma^k \in \mathbb{R}_{+}^2$ its standard deviations, and $\rho^k \in (-1,1)$ its correlation.
In our location model, a single GMM describes the distribution $p(y_t \mid I)$. 
In our stroke model, the LSTM outputs one GMM at each timestep, describing $p(\Delta_t | \Delta_{1:t-1}, y_t, I)$.

\subsubsection{Training.}
Our Full NS model provides a density function which can be used to score the log-likelihood for any character drawing.
We train the model to maximize the log-likelihood (minimize log-loss) of the training set drawings, using mini-batch gradient descent with a batch size of 200 and the Adam update rule.

\section{Alternative Models}
\begin{figure}[t]
    \centering
    \includegraphics[width=\hsize]{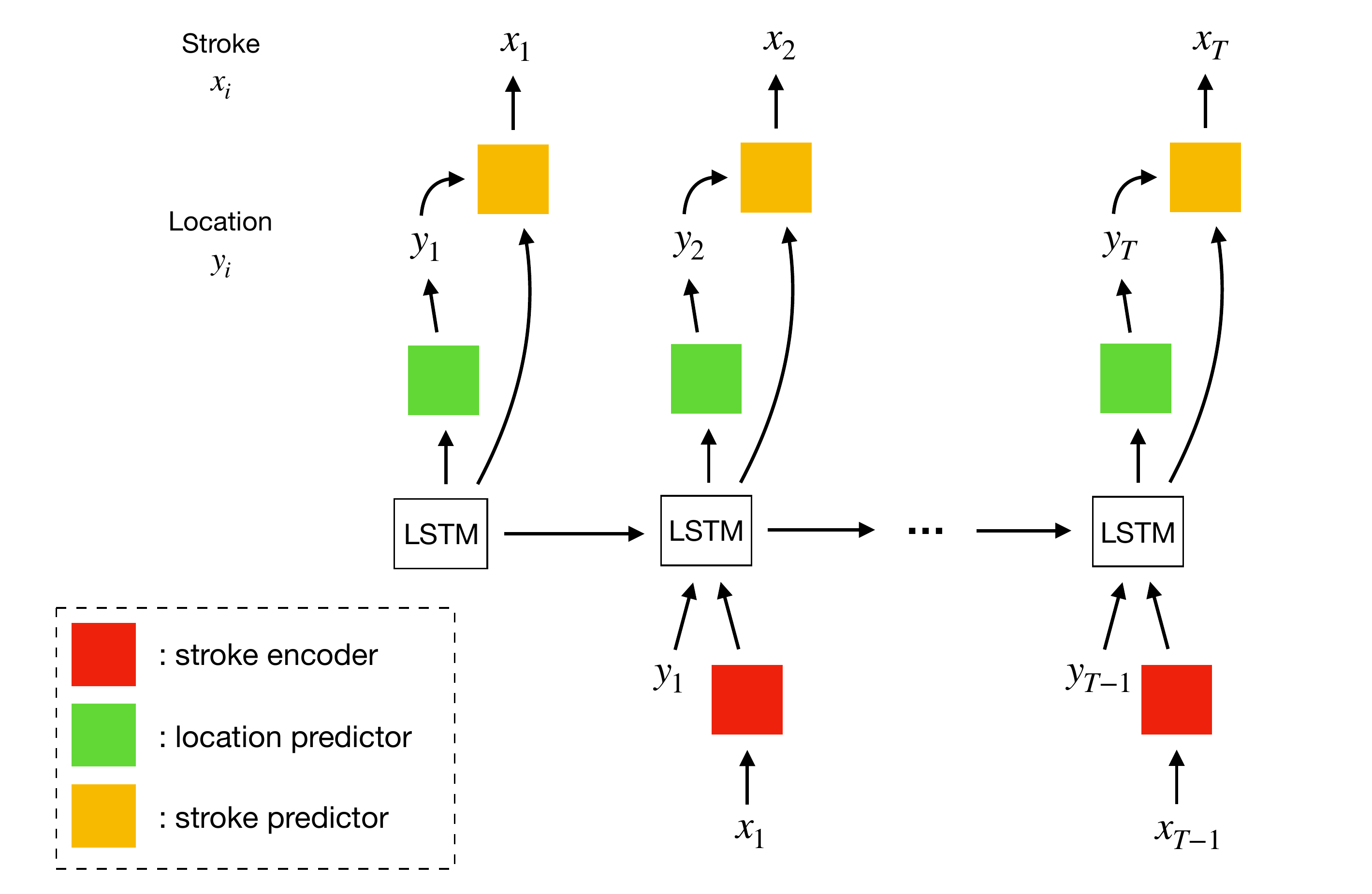}
	\bcaption{Hierarchical LSTM model.}{ 
	The model samples characters one stroke at a time, using a character-level LSTM as a memory state. 
	At each time, the model samples a starting location for the next stroke from a location predictor (MLP), and a stroke trajectory from the stroke predictor (LSTM). 
	These samples are then fed to the model as inputs for the next time, with the location fed directly and the trajectory processed by a stroke encoder (bi-directional LSTM).
	}
    \label{fig:arch_hlstm}
\end{figure}
In addition to our Full NS model, we explored two alternative models with more generic neural network architectures.
In each alternative, we lesioned key structural ingredients of the Full NS model, hoping to test the importance of these ingredients to model performance.

\subsubsection{Hierarchical LSTM.} 
As one alternative neural model, we explored a hierarchical recurrent architecture \citep{Sorodini2015,Ling2016,Chung2017}, which we denote ``Hierarchical LSTM" (H-LSTM).
Like our Full NS architecture, the H-LSTM model is trained on causal data demonstrating how people actually produce drawings of characters.
In addition, it models the compositional structure of characters by separating them into explicit stroke parts, which defines the hierarchy in the hierarchical LSTM.
Unlike our Full NS model, however, the H-LSTM has no renderer and thus lacks any explicit causal  knowledge of how motor actions become raw images of inked characters.
Instead, information about the previous strokes is written to memory via recurrent connections and gating mechanisms. 
These transformations can propagate arbitrary correlations, and they must be learned entirely from the data.

Specifically, at each time step $t$, the previous stroke $x_{t-1}$ is read by a \textit{stroke encoder} $f_{\text{enc}}$, a bi-directional LSTM that processes the stroke and returns a fixed-length vector (red box in Fig. \ref{fig:arch_hlstm}). This vector is then passed as an input to the character LSTM along with previous location $y_{t-1}$ and previous hidden state $h_{t-1}$:
\begin{align*}
    h_t = f_{\text{LSTM}}(y_{t-1}, f_{\text{enc}}(x_{t-1}), h_{t-1}).
\end{align*}
The new hidden state $h_t$ is then fed to the \textit{location model} $p(y_t \mid h_t)$, a multi-layer perceptron that outputs a GMM distribution for the next stroke's starting location $y_t$ (green box in Fig. \ref{fig:arch_hlstm}).
The location is sampled from this distribution and passed as an input along with $h_t$ to the \textit{stroke model} $p(x_t \mid h_t, y_t)$, an LSTM that samples a stroke trajectory one offset at a time with GMM outputs (yellow box in Fig. \ref{fig:arch_hlstm}):
\begin{align*}
    y_t &\sim p(y_t \mid h_t) \\
    x_t &\sim p(x_t \mid h_t, y_t).
\end{align*}

\subsubsection{Baseline LSTM.} 
A second alternative is even less structured and represents the most purely \textit{statistical} architecture we examined. 
For this model, we explored a naive unrolled LSTM, denoted ``Baseline."
This model is a reproduction of the unconditional version of Sketch-RNN \citep[][Sec 3.3]{Ha2018}.
Similar to Full NS and H-LSTM, the Baseline LSTM is trained on causal data demonstrating the process of producing characters; however, the architecture does not explicitly take compositional structure into account. 
Instead, it uses a single RNN to model a character as one long sequence of pen actions with stroke breaks.

Following Sketch-RNN, we expand the binary pen state variable $v_t \in \{0,1\}$ from \cite{Graves2013} to a ternary variable $v_t \in \{0,1,2\}$ to handle multi-stroke drawings. Value 0 indicates that we are continuing the current stroke, 1 that we are ending the current stroke and starting a new one, and 2 that we are ending the drawing.
The initial hidden and cell states of the LSTM are set to zero, and at each time step $t$, the previous offset $\Delta_{t-1}$, previous pen state $v_{t-1}$, and previous hidden state $h_{t-1}$ are fed as inputs to the LSTM, which outputs new hidden state $h_t$: 
\begin{align*}
    h_t = f_{\text{LSTM}}(\Delta_{t-1}, v_{t-1}, h_{t-1}).
\end{align*}
An output layer receives $h_t$ and returns a categorical distribution for next pen state $v_t$, and a GMM for next offset $\Delta_t$:
\begin{align*}
    \theta_v &= f_v(h_t), \quad v_t \sim p(v_t \mid \theta_v) \\
    \theta_{\Delta} &= f_{\Delta}(h_t), \quad \Delta_t \sim p(\Delta_t \mid \theta_{\Delta}).
\end{align*}

\section{Experiments}
We evaluated the creative generalizations of our 3 models using both quantitative and qualitative analyses. 
Each of our models estimates a probability density function for characters from training examples. 
This density function can be used to compute likelihoods for held-out characters and to generate new character samples. 
A generative model for characters that exhibits creative generalization should produce high likelihood scores for novel character concepts from held-out classes. 
In addition, the model should generate new characters that are sufficiently dissimilar from the training examples, but that are structurally consistent with ground truth. 
In our quantitative analysis, we tested our models for their likelihood performance on novel character classes using a rigorous set of experiments with different train/test splits. 
In our qualitative analysis, we inspected the character samples, comparing with BPL, ground truth concepts, and nearest neighbors from the training set.

\begin{table}[t]
    \scriptsize
    \centering
    \begin{tabular}{l|l|l|l|l|l|l|l} 
        & \multicolumn{3}{c|}{Alphabet Splits} & \multicolumn{3}{c|}{Character Splits}
        & Holdout \\
        \hline
        Model  & split1 & split2 & split3 & split1 & split2 & split3 &  - \\
        \hline
        Full NS      &\textbf{13.77} &\textbf{14.18} &\textbf{17.53} &12.35          &\textbf{12.59} &12.57          & \textbf{19.51}\\
        H-LSTM      &14.37          &14.56          &17.71          &12.24          &12.80          &12.51          & 20.16\\
        Baseline    &14.32          &14.42          &17.71          &\textbf{12.20} &12.77          &\textbf{12.39} & 19.66\\
    \end{tabular}
    \bcaption{Test losses from our 3 models.}{
    Losses indicate the average negative log-likelihood per test character (lower is better).
    In our “alphabet splits” task, we divide the background set into train/test splits such that the model must generalize to new characters from novel alphabets. 
    In our ``character splits” task, we divide the background set such that the model must generalize to new characters from familiar alphabets. 
    In our ``holdout” task, we provide the entire background set for training and use the held-out evaluation set--which contains new characters from novel alphabets--for testing.
    }
    \label{tab:results_loglike} 
\end{table}
\begin{figure*}[t]
    \centering
    \includegraphics[width=\hsize]{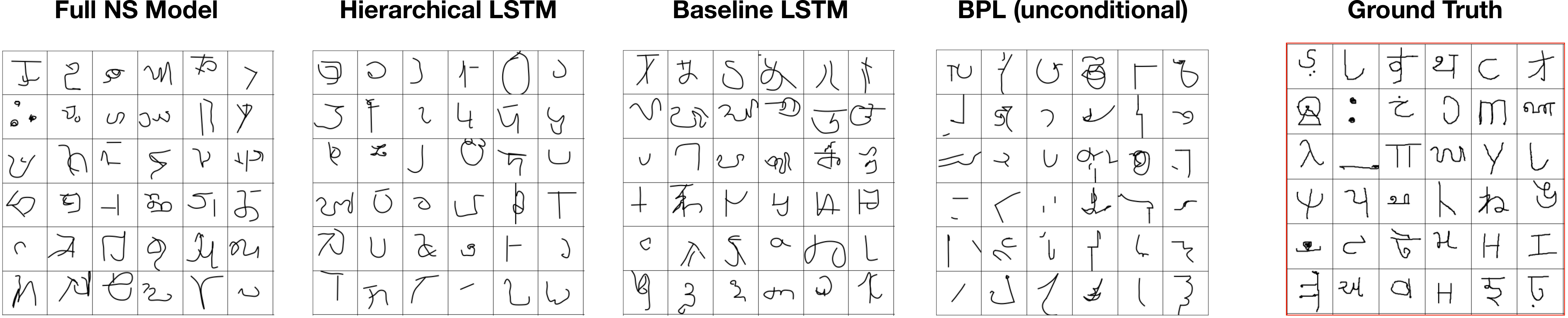}
    \bcaption{Character sample comparison.}{Characters generated by our Full NS, H-LSTM and Baseline LSTM models are shown side-by-side, along with samples from the BPL forward model\textsuperscript{2} as well as ground truth characters from Omniglot.}
    \label{fig:sample_comparison}
\end{figure*}

\subsection{Evaluation on Held-Out Concepts}
\subsubsection{Methods.}
In our quantitative analysis, we evaluated our models for two different forms of likelihood generalization, corresponding to different train/test splits. 
In the first generalization task, denoted ``character splits," we asked whether our models could generalize to new character classes from \textit{familiar alphabets}. 
We created 3 train/test splits from the Omniglot background set, sampling 80\% of characters per alphabet for train and 20\% for test. 
In our second task, denoted ``alphabet splits," we asked whether our models could generalize to new character classes from \textit{novel alphabets}. 
We again sampled 3 train/test splits of size 80-20, this time splitting by alphabet.
In both the ``character splits" and ``alphabet splits" tasks, we explored multiple hyperparameter configurations for our models, varying parameters such as the number of hidden layers, number of units per layer, and dropout probability.
\footnote{For details about hyperparameters, see Appendix A.}
Average validation loss across splits was used to select the best configuration for each model in each task.
We then took our best configurations in each task and reported their validation losses on all 3 splits.

As a final quantitative analysis, we tested our models on one additional task that extends the ``alphabet splits" task.
Our motivation was to provide a more rigorous analysis using a completely withheld test set as per standard practice in machine learning evaluations.
We re-trained our best configurations of each model on the entire background set, using the hyperparameters selected from our ``alphabet splits" task. 
We then reported losses on the evaluation set, which contains character drawings from 10 completely novel alphabets.

\subsubsection{Results.}
Results from the cross-validation splits are shown in Table \ref{tab:results_loglike}, ``Alphabet Splits" and ``Character Splits."
In our alphabet splits, the Full NS model consistently outperformed the alternatives, exhibiting the best generalization performance in each of the 3 splits.
Thus, our neuro-symbolic architecture appears best equipped to capture overarching principles in handwriting concepts that generalize far outside of the training examples.

In our character splits task, the Baseline LSTM exhibited best performance in 2 out of 3 splits, and the Full NS model in 1 of 3.
The character splits present a much easier generalization task, where exemplar-based learning could offer a suitable alternative to learning general structural principles.
Interestingly, the selected hyperparameter configuration for our Full NS model remained constant across the ``alphabet" and ``character" split tasks, whereas the configuration changed for both the Baseline and H-LSTM models.

Results for each model on the held-out set of characters are shown in Table \ref{tab:results_loglike}, ``Holdout."
Similarly to the ``alphabets" task, our Full NS model outperforms both alternative models on the holdout set, providing further support that this architecture learns the best general model of these simple visual concepts.
A paired t-test reveals the Full NS model has reliably better loss per example than the next-best model (Baseline; t(5531) = 3.094; $p < 0.002$).

\begin{figure}[t]
    \centering
    \includegraphics[width=\hsize]{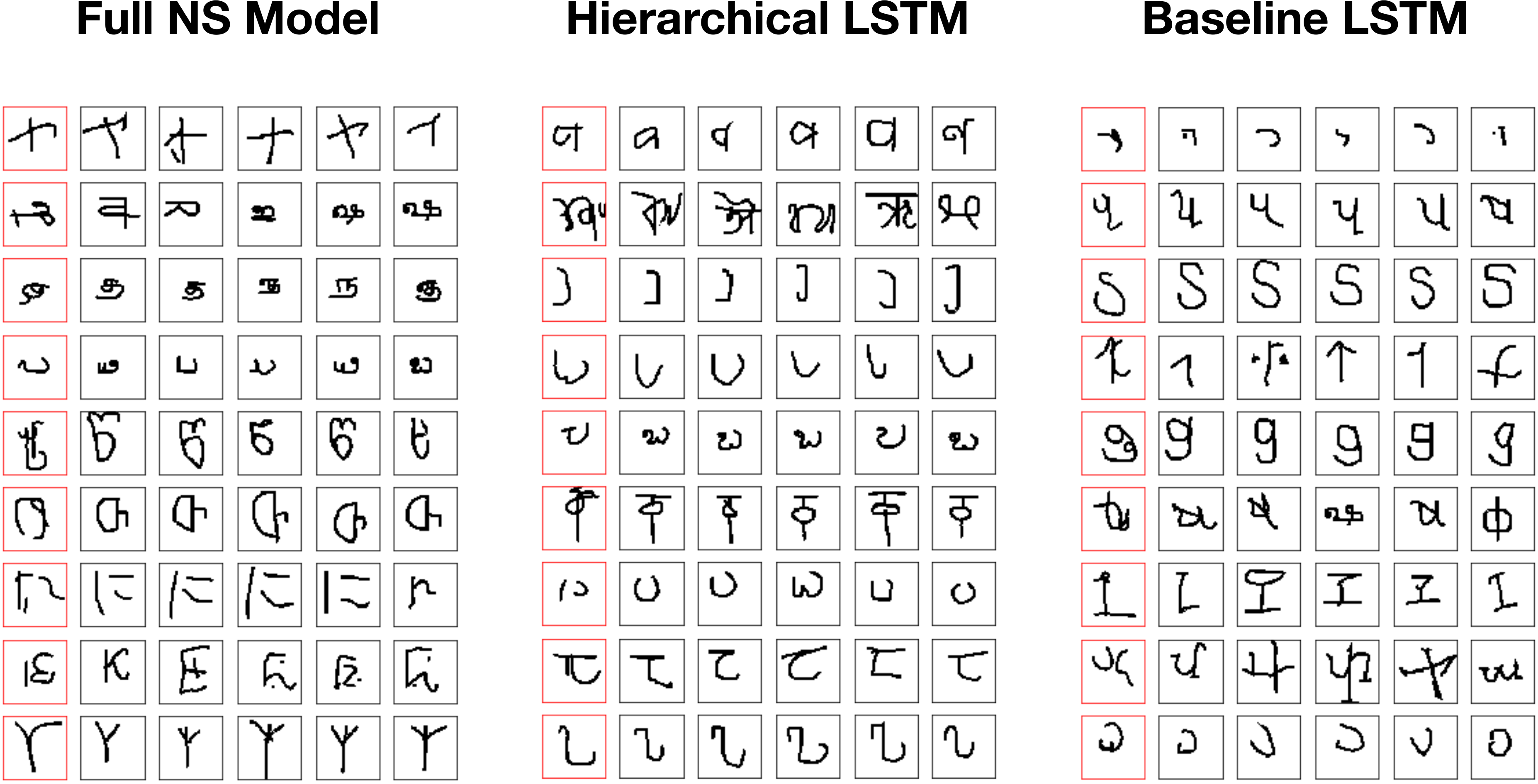}
	\bcaption{Novelty of character samples.}{
	Character drawings sampled from each model were compared to their 5 nearest neighbors from the training set. Each row corresponds to one character sample from the model. The red box indicates the model sample, and the 5 nearest neighbors are shown in the succeeding columns.
	}
    \label{fig:nearest_neighbors}
\end{figure}

\subsection{Generating New Concepts}
\subsubsection{Methods.}
In our qualitative analysis, we analyzed the 3 neural network models on their ability to produce novel visual concepts.
We took our trained models from the previous experiment and sampled 36 characters from each model, following the model's causal generative procedure. 
In addition, we sampled 36 characters from the BPL character prior, and we selected 36 ``ground truth" characters from Omniglot at random. 
Samples were then compared visually side-by-side.

As an additional qualitative analysis, we compared character samples from each model for their similarity to the training examples.
Although the complexity and structural coherence of generated characters are important criteria, these observations alone provide insufficient evidence for a human-like generative process; a model that memorizes the training examples might produce samples with structural coherence and rich variations, but such a model does not account for the flexible ways that humans generate new concepts.
In our second analysis, we took the character samples from our models and found the 5 most-similar training characters for each, using cosine distance in the last hidden layer of a CNN classifier as a metric space for perceptual similarity. 
The CNN was trained to classify characters from the Omniglot background set, a 964-way classification task.

\subsubsection{Results.} 
Fig. \ref{fig:sample_comparison} shows samples from each of our three models, as well as from the BPL forward model\footnote{BPL character samples have been centered for better visual appearance; the actual samples often protrude outside of the image window. A more complex non-parametric BPL model was used in the visual Turing tests in \citet{Lake2015} that has explicit re-use of character parts. Those samples were also centered.} and from the Omniglot data (ground truth).
Compared to BPL, the neural-enhanced models capture more correlational structure and character complexity.
For instance, the Full NS model propagates stylistic and structural consistency across three strokes to form a Braille-like character, as shown by the sample in column 1, row 2.
Fig. \ref{fig:nearest_neighbors} shows a handful of character samples produced by each neural model plotted alongside their five nearest neighbors from the Omniglot training set. Unlike the log-likelihood results, comparing models in this fashion is subjective; nevertheless the H-LSTM and Baseline LSTM produce more characters that closely mimic the nearest training examples (7/9 by our eyes). In contrast, our Full NS model produces only a few (3/9) characters that are close mirrors of training examples, suggesting that it can generalize further from the training observations.

\begin{figure}[t]
    \centering
    \includegraphics[width=\hsize]{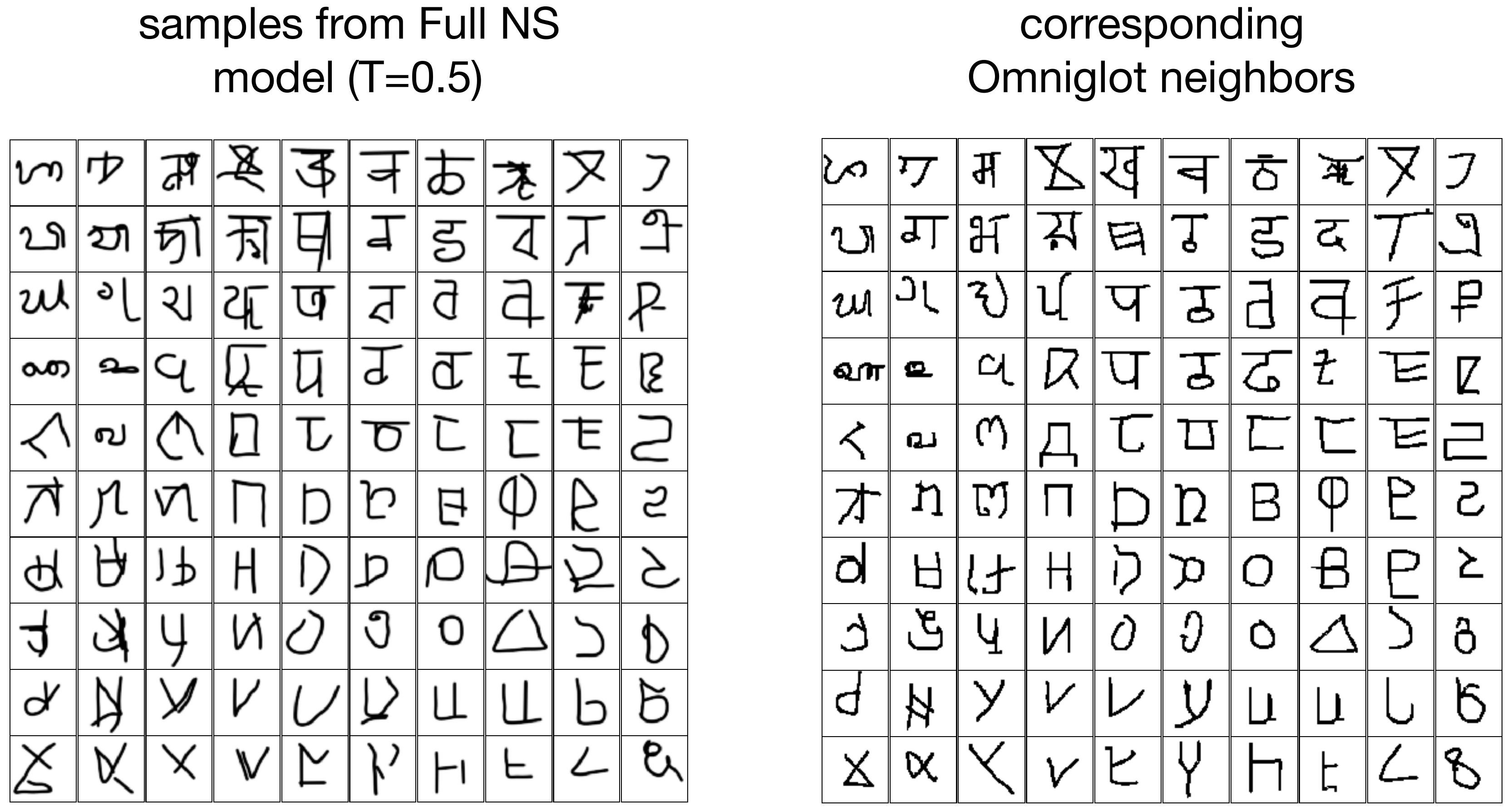}
	\bcaption{Topologically-Organized character samples and their nearest Omniglot neighbors.}{
	We drew 100 character samples from our Full NS model and organized them into a 10x10 grid such that neighboring characters have similar drawing styles (left). 
	We then found the ``nearest neighbor" of each sample from the Omniglot character dataset and organized the neighbors into a corresponding 10x10 grid (right).
	}
    \label{fig:topographical_samples}
\end{figure}

To get an idea of the different character styles produced by our Full NS model, we sampled 100 characters from the model and organized them into a 10x10 grid such that neighboring characters have high perceptual similarity (Fig. \ref{fig:topographical_samples}, left).
Characters were sampled at a lower level of stochasticity, using the temperature parameter proposed by \cite{Ha2018} to modify the entropy of the mixture density outputs (we used $T=0.5$).
The model produces characters in multiple distinct styles, with some having more angular, line-based structure and others relying on complex curves.
In Fig. \ref{fig:topographical_samples} (right), we plotted the most-similar Omniglot character for each sample in a corresponding grid.
In many cases, samples from the model have a distinct style and are visually dissimilar from their nearest Omniglot neighbor.

\section{Conclusion}
We presented a new neuro-symbolic generative model of simple visual concepts. 
Our model successfully captures compositional and causal structure in handwritten character concepts, forming a representation that generalizes to new concepts. 
We tested our model by comparing its likelihood scores on a holdout set of novel characters, finding that it consistently outperforms two generic neural network alternatives when the test characters deviate significantly from the training examples. 
Furthermore, our generative model produces new character concepts with richer variations than simple parametric models, yet that remain structurally coherent and visually consistent with human productions.

Neuro-symbolic models offer a promising set of tools to express the rich background knowledge that enables creative generation. 
These models can explain the nonparametric correlation structure embodied in conceptual knowledge while maintaining important inductive biases to account for the structured ways that people generate new concepts.
We believe that models of this kind will be useful to explain a variety of human imaginative behaviors, such as when a chef creates the new recipe ``pea guacamole." In future work, we'd like to explore applications of neuro-symbolic models to other types of concepts with varying complexity.

\subsection{Acknowledgements}
We thank Maxwell Nye, Josh Tenenbaum, Tuan-Anh Le, and Jay McClelland for helpful discussions regarding this work. Reuben Feinman is supported by a Google PhD Fellowship.

\bibliographystyle{apacite}

\setlength{\bibleftmargin}{.125in}
\setlength{\bibindent}{-\bibleftmargin}

\bibliography{references}

\newpage
\appendix
\section{Appendix}

\subsection{A. Model Hyperparameters}
Here we review the hyperparameters (HPs) used for each of our models, indicating which HPs were fixed and which were tuned. All neural networks with GMM output layers use 20 mixture components.

\subsubsection{Full NS.} The Full NS model has 3 submodules: a \textit{location} model, a \textit{stroke} model, and a \textit{termination} model. 
Each submodule uses a distinct CNN, and each receives an image canvas of size (28,28). 
The \textit{location} and \textit{termination} models--which return outputs for a single time step--each use a feed-forward CNN architecture inspired by \cite{Vinyals2016}. 
The CNNs consist of a stack of 4 blocks, with each block $i$ including a 3x3 convolution with $K_i$ filters, batch normalization, nonlinear activation $f$, 2x2 max-pooling, and dropout with rate $p$. 
These blocks are followed by a single fully-connected layer with $D$ units, activation $f$ and dropout $p$.
Hyperparameters $\{K_i\}$, $f$, $p$ and $D$ were selected from tuning. 
The \textit{stroke} model uses a modified CNN architecture without spatial pooling, designed to convey high-resolution spatial information for visual attention. 
The CNN returns a feature map of size (64, 14, 14), which is then passed to an LSTM. The LSTM predicts the spline trajectory of the next stroke one offset at a time, attending to different parts of the feature map at each step. The HPs of the CNN were fixed, but the HPs of the LSTM were tuned, including the number of LSTM layers and number of units per layer.

\subsubsection{Hierarchical LSTM.}
The Hierarchical LSTM model has a character-level LSTM backbone and 3 submodules: a \textit{stroke encoder} (BiLSTM), a \textit{location model} (MLP), and a \textit{stroke model} (LSTM). 
The number of LSTM layers, number of units per layer and dropout rate in the character-level LSTM were selected from tuning, but HPs of all submodules were fixed.
The \textit{stroke encoder} is a bidirectional LSTM with a single layer of 256 units. 
It outputs a fixed-length vector representation of the previous stroke, which is fed to the character LSTM as input. 
The \textit{location model} is a 2-layer MLP that receives the current hidden state of the character LSTM and outputs a GMM for the next stroke's starting location. 
The \textit{stroke model} is an LSTM with a single layer of 256 units and outputs a GMM at each time step for the next spline offset.

\subsubsection{Baseline LSTM.}
The Baseline LSTM is a single module. It has $L$ LSTM layers, each with $K$ units and dropout rate $p$. The values of $L$, $K$ and $p$ were selected from tuning.

\subsection{B. Samples with stroke decomposition}
\begin{figure}[t]
    \centering
    \includegraphics[width=\hsize]{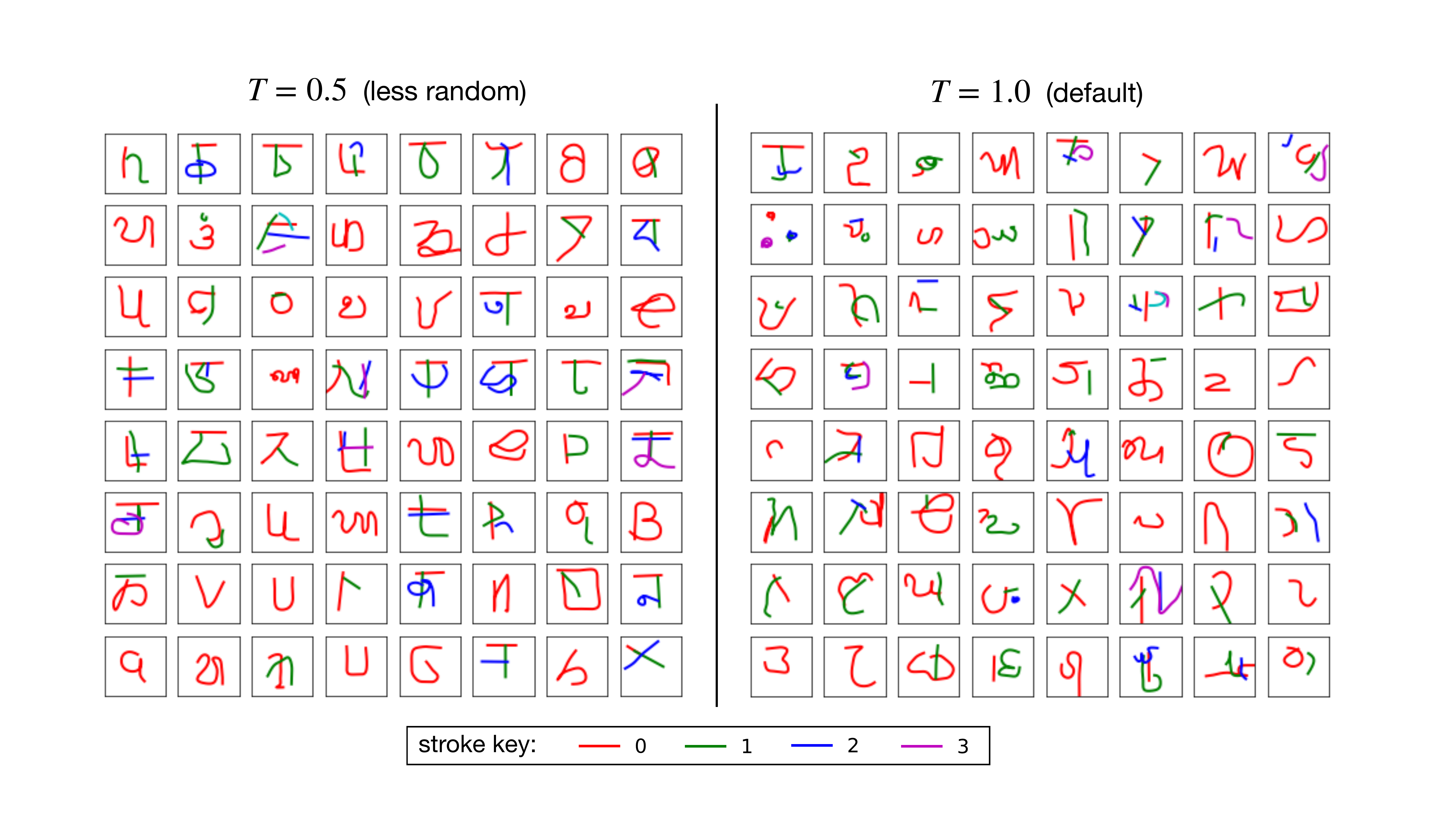}
	\bcaption{Samples with stroke decomposition.}{
	Character samples produced by our Full NS model are shown with stroke decompositions. 
	Samples were produced at two temperature settings \cite[Eq.8]{Ha2018}, using $T=1.0$ and $T=0.5$.
	}
    \label{fig:canvas_samples}
\end{figure}
In Fig. \ref{fig:canvas_samples}, we show a larger collection of characters from our Full NS model, using color coding to convey the stroke composition of each sample.
We produced character samples at two different levels of stochasticity, using a temperature parameter to modify the entropy of the mixture density outputs \cite[Eq.8]{Ha2018}. Samples are shown for temperature settings $T=1.0$ and $T=0.5$.

\end{document}